# A Synergistic Approach for Recovering Occlusion-Free Textured 3D Maps of Urban Facades from Heterogeneous Cartographic Data

Regular Paper

Karim Hammoudi[1,2,*], Fadi Dornaika[3,4], Bahman Soheilian[2], Bruno Vallet[2], John McDonald[1] and Nicolas Paparoditis[2]

1 Department of Computer Science, National University of Ireland Maynooth, Maynooth, Co. Kildare, Ireland
2 Université Paris-Est, Institut Géographique National, Laboratoire MATIS, Saint-Mandé, France
3 Department of Computer Science & Artificial Intelligence, University of the Basque Country, San Sebastián, Spain
4 IKERBASQUE, Basque Foundation for Science, Bilbao, Spain
* Corresponding author E-mail: karim.hammoudi@cs.nuim.ie

Abstract In this paper we present a practical approach for generating an occlusion-free textured 3D map of urban facades by the synergistic use of terrestrial images, 3D point clouds and area-based information. Particularly in dense urban environments, the high presence of urban objects in front of the facades causes significant difficulties for several stages in computational building modeling. Major challenges lie on the one hand in extracting complete 3D facade quadrilateral delimitations and on the other hand in generating occlusion-free facade textures. For these reasons, we describe a straightforward approach for completing and recovering facade geometry and textures by exploiting the data complementarity of terrestrial multi-source imagery and area-based information.

## 1. Introduction

Nowadays, developments of Mobile Mapping Systems equipped with direct georeferencing devices are revolutionising the mapping world (e.g., [1, 2]). These MMSs acquire rich and massive multi-source data at street level that can be useful to break down some barriers for significantly enhancing the quality of urban facade maps at wide scale. We describe below some current mapping constraints that can potentially be overcome.

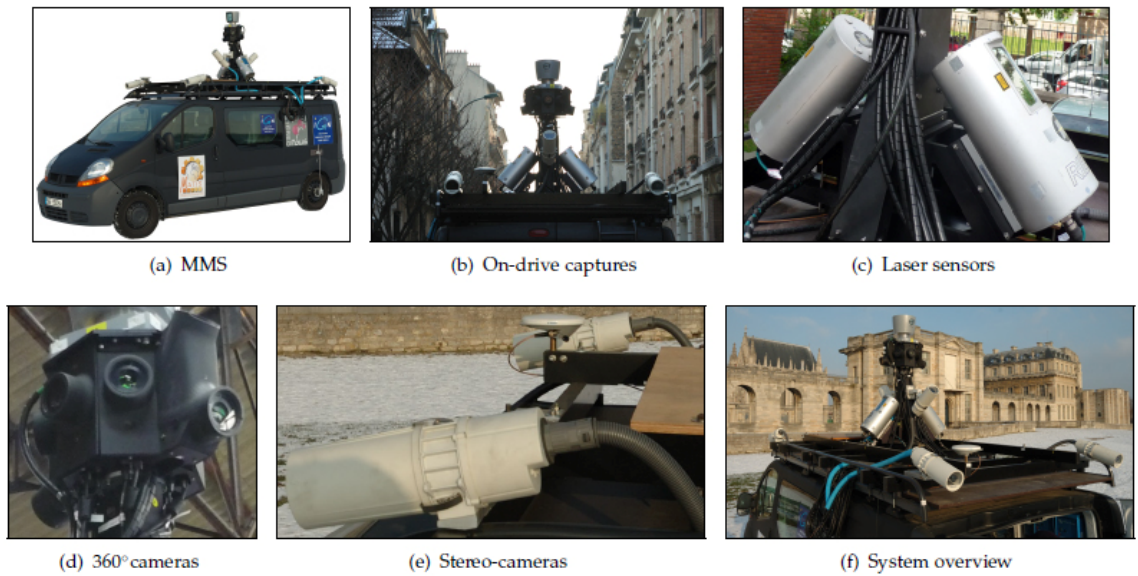

**Figure 1.** Mobile Mapping System, protocol and devices of acquisition.

**Aerial-based Mapping Constraints** A large number of approaches are proposed in the literature for generating 3D building models by using aerial data (e.g., [3]). These 3D building models are usually textured at roof level by using aerial images and sometimes textured at facade level by exploiting oblique aerial images (e.g., [4]). Hence 3D complete building models can be reconstructed and textured. Alas, these building models are often inappropriate for terrestrial urban walk-through visualization. Foremost, the aerial-based facade textures are often very poor in quality (low resolution, high perspective deformations) due to the distance and the angle of the points of acquisition (see [1]). Furthermore, the generated 3D aerial-based delimitations are very inaccurate at facade level due to the fact that many ground/wall and wall/roof junctions of facades are not visible in aerial views. Thus, the ground/wall and wall/roof junctions of facades are frequently confused with gutter limits vertically extruded and their estimated heights and planimetric locations can deviate of several decimetres by comparison with the reality.

**Terrestrial-based Mapping Constraints** Calibrated street images with a high resolution and accurate georeferencing can be collected by MMSs (e.g., see Fig.1(a)). Subsets of facade images are often automatically mapped onto 3D facade quadrilaterals of aerial-based building models (i.e., facade texturing). This image-to-model matching lacks consistency and registration problems appear in part due to high differences of scale in the processed data. Hence, resulting terrestrial-based facade textures can be non-representative of true facades (e.g., partial representation of facade walls or representation of external objects such as roof portions) due to ill-fitting aerial-based delimitations. These problems also strongly appear in textured building models that are semi-manually generated by designers or architects (high consumers of building models) and where demonstrative virtual platforms are created for presenting urban planning projects. Moreover, a critical and widespread problem of these urban facade textures lies in the presence of strong occlusions caused by the urban street objects (unpredictable occlusions). In many cases, this problem is not addressed and acquired frontal facade images are directly mapped onto the generated or created models in spite of the presence of wide visible occlusions (e.g., [5]). Furthermore, this problem affects a huge amount of surrounding processing and stages in computational building modeling. For example, some powerful image-based segmentation approaches are robust to partial facade occlusion (e.g., using grammar shapes [6]). However they are not robust to large scale occlusions.

For solving part of these problems that are particularly common in dense urban environments, we propose a synergistic approach in which heterogeneous cartographic data greatly take advantage of each-others strengths for generating accurate 3D facade quadrilaterals (e.g., [1]) as well as occlusion-free facade textures that correctly match at this level of representation (planar models).

The produced textured facade models can be useful information for many applications such as virtual navigation, autonomous navigation (e.g., [7, 8]) and intelligent transportation systems (e.g., [9]). Indeed, vision-based map-less navigation can exploit the set of textured facades in order to match the on-board acquired images with the corresponding georeferenced facades. This results in a positioning system that is only based on images (e.g., [10]).

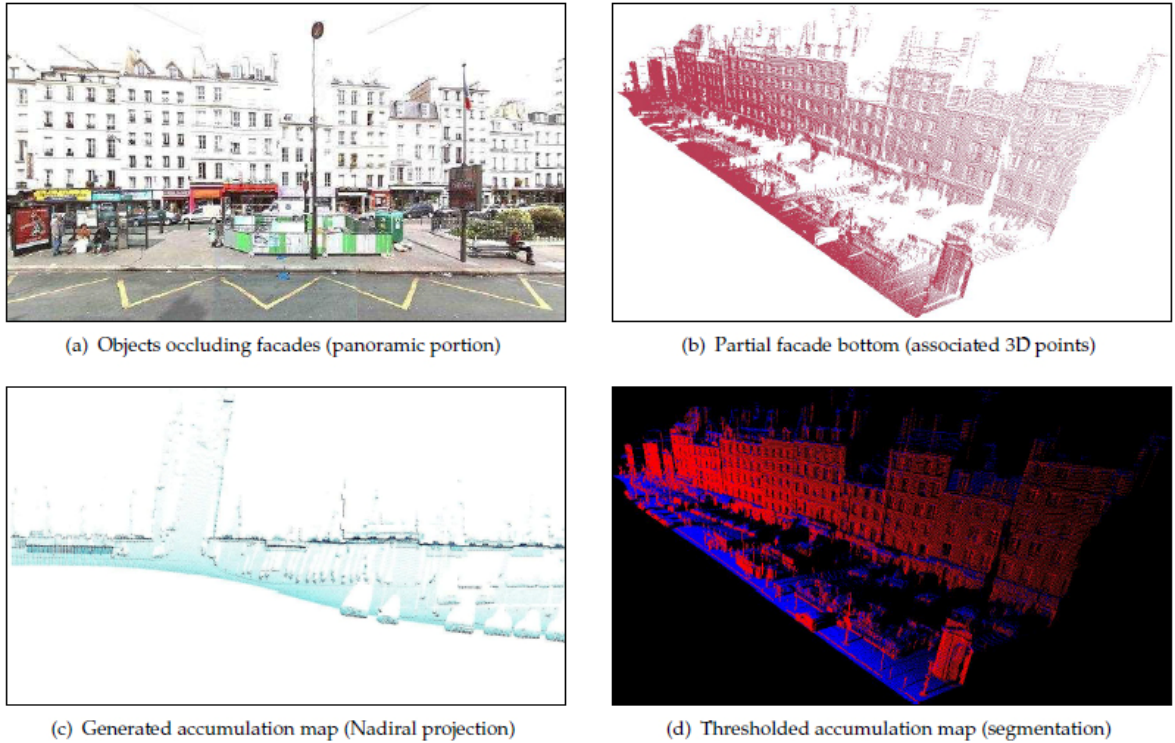

**Figure 2.** Results of the primary street point cloud segmentation.

## 2. Operating Mobile Mapping System

Here, the MMS (see Fig.1(a)) is equipped with two high precision 2D laser sensors RIEGL LMS-Q120i (see Fig.1(c)). These laser sensors are positioned at the mast of the vehicle (see Fig.1(b)) and respectively oriented to each street side in direction of the facades. Their beam plane is perpendicular to the vehicle trajectory. The system allows us to carry out up to 10000 measurements per second and the beam vertically sweeps with an opening of $\sphericalangle \alpha \simeq 80°$ (−20° to 60° with respect to the horizontal). This vertical sweeping provides a frame of points (i.e., range or profile). The angular resolution was configured to 201 points per frame. The nominal accuracy of the laser-based measurements is approximately 3*cm* at 150*m*. The laser sampling is approximately one point each 5*cm* to 10*cm* in planimetry and altimetry when the vehicule moves at 30*km/h*. These sensors are used for the geometric modeling of building facades. The MMS masthead (Fig.1(b)) is composed of eight cameras that are oriented in cardinal and ordinal directions in order to cover the scene at 360° (see Fig.1(d)). Two additional cameras are laterally located with an angular sloping $\sphericalangle \beta \simeq 45°$ in direction of the top of the facade. The vehicle is also equipped with two stereoscopic acquisition systems located in the front and in the back of the vehicle and oriented in direction of the road (see Fig.1(e) and Fig.1(b)). All these cameras are Full HD AVT Pike 210C digital cameras that acquire 16*bit RGB* images in 1920 × 1080, namely around $20\overline{M}$ of pixels. The Ground Sample Distance of these images can reach up to $\simeq 5cm$ *GSD*. The cadence of acquisition is one capture each 4.5*m* or each 10*s* (grouped captures). These cameras are used in the texturing of facade models. Besides, positioning instruments are used for georeferencing all the collected data. An overall view of image and laser sensors mentioned above is exhibited in Figure 1(f).

## 3. Proposed Approach

*3.1 Primary Street Point Cloud Segmentation*

*3.1.1 Accumulation Map Generation (Euclidean)*

The verticalness of the street facade walls and other vertical urban objects is exploited. A 2D accumulation map is generated by projecting the 3D point cloud vertically into an horizontal regular grid, i.e. map of accumulation noted $A_{map}$ (see Fig.2(c)). The grid step $\left( \Delta_{(u,v)} = 5cm \times 5cm \right)$ is roughly selected in a value order close to the laser sensor resolution. Usually, it reaches up to hundreds of points per square meter in the horizontal and vertical planes. For each 3D point (**X**, **Y**, **Z**) of the raw street point cloud (Fig.2(b)), the associated 2D planimetric coordinates noted (**u**, **v**) produce a vote in a cell of the map. The score of a cell is noted $S_{(u,v)}$. Since the relation between 3D points and 2D cells is kept, a raster data $A_{map}$ complementary to the 3D point cloud is generated.

| Geometric Characteristic / Semantic Attribute | **VerticalSet** $H_\perp$ (Hyper-points) | **SurfaceSet** $H_\perp^c$ (Complementary) |
|---|---|---|
| **Dominant urban superstructure** | facade (wall & window) | ground surface roof surface |
| **Street road microstructure** (Facade foreground) | parked vehicle, barrier, bus shelter, kiosk, phone booth, urban post, street light, bench, road sign, urban bean, tree | - |
| **Roof microstructure** (Facade background) | dormer, chimney, antenna | - |

**Table 1.** Types of identified urban street objects resulting from the primary segmentation of the produced accumulation map $A_{map}$ (e.g., observations of feature points (sometimes partial) shown in Fig.2(c) and Fig.2(d)). At least 15 types of frequent urban objects can be recognized (objects also recognized in other related street datasets are mentioned). Notably, the identified objects are classified into categories (geometric vs. semantic). These objects are regrouped into 4 categories. In this study, the interest particularly lies in the category including facades.

*3.1.2 Map-based Segmentation into Vertical and Surface Clusters*

A global density threshold is applied to the cells (**u**, **v**) of the generated map $A_{map}$ (see Fig.2(d)). The 3D points that vote in cells with a score $S_{(u,v)} > 1$ are labelled as potential vertical objects and are denoted hyper-points[1]. This set of hyper-points characterizes the majority of facades and street road microstructures (streetlights, posts, road signs, tree trunks) since their cells have a high vote. If visible, the roof superstructures are also contained into the obtained cluster (e.g., chimneys, antennas). The remaining 3D points (cells with a score $S = 1$) are labelled as potential non-vertical flat surfaces. The associated cluster constitutes the majority of ground and roof plane portions. Some urban objects identified from the labelled points are respectively summarized in red and blue columns of Tab.(1). Besides, we note $H_\perp = \{(X,Y,Z) \subset (u,v) \mid S_{(u,v)} > 1\}$ the subset of hyper-points that can be exploited for the potential extraction of various vertical urban objects in 3D. The set of complementary points (labelled non-vertical) is noted $H_\perp^c$. Some urban objects identified from the detected points are categorized in columns of Tab.(1). Next stages aim at extracting the subset of hyperpoints (red column) corresponding to 3D facade points (category vertical and dominant superstructure) and then estimating the associated facade quadrilaterals.

*3.2 Extracting Clusters and Estimating 3D Quadrilateral Delimitations of Urban Facades*

Many cities and large capitals throughout the world detain standard building maps that have been produced by surveyors. These maps that are often composed of georeferenced facade segments derived from measurements carried out from laser-based tachymeters. Notably, the **global coherence** of such maps is attested across their numerous uses in current aerial-based and wide-scale building modeling pipelines as well as in a huge amount of surrounding urban applications (e.g. taxations based on occupation of dwelling surfaces). Besides, recent evaluations demonstrate wide-scale compatibility between mobile laser scanning data and conventional cadastral maps for generating highly detailed 3D building maps (e.g., [11]). In our case, it is utilised in the estimation and delimitation of planar facade models.

*3.2.1 Focusing on Facade Regions by Exploiting Existing Digital Cadastral Maps*

Here, a standard **DCM** (Digital Cadastral Map noted here $C_{map}$) coming from national GIS databases handled by a mapping agency is employed for pre-determining the initial location of facade extremities in planimetry. This map is composed of 2D facade segment lines which are also georeferenced into 3D (see sample of segments in Fig.3(b)). Generally, it is observed that this kind of map is heterogeneous in accuracy (depending on the surveyor proficiency) but often respects a bounded inaccuracy interval of some decimetres according to the reality. Hence, we can hope that the facade segments given by the map can provide relevant **ROIs** (Regions Of Interest) in order to detect close facade clusters and to estimate accurate facade delimitations from the subset of points $H_\perp$ previously computed (see Fig.3(a)). In resume, we propose to employ an available existing $C_{map}$ when the known associated characteristics respect some quality criteria (e.g., accurate georeferencing, accurate and complete geometry). Besides, it is worth mentioning that these maps were in many cases extremely expensive to produce in reason of the high level of human interventions. Although expensive to produce, the $C_{map}$s are **relatively available** since often become low-cost or free (e.g. in the public domain) for numerous cities. Furthermore, these maps are connected to several databases (e.g., address, historic). This richness of information undoubtedly results from a great regrouping effort of urban knowledge. It appears interesting to establish a link between current facades of the $C_{map}$ and their homologous in the next generation of 3D facade maps (e.g., updated facade maps) for example for directly transferring semantic information or modeling the urban morphogenesis. An idea here consists of **"recycling a cadastral map"** for producing a novel accurate 3D facade map by coupling our $C_{map}$ with MMS-based laser data.

---

[1] A hyper-point is defined as a point voting in a cell having a high score. Here, the notion is derived from [3] which defined a hyper-point as a series of 3D points having the same x-y coordinates but different z values.

*3.2.2 Extracting Disjoint Facade Clusters by Fusing $A_{map}$ and $C_{map}$ Information*

The $C_{map}$ (Fig.3(b)) is employed for initializing planimetric locations of the searched corresponding facade points. The previously segmented laser dataset (Fig.3(a)) and the exploited $C_{map}$ are both georeferenced and expressed in the same world coordinate system. We assume that the misalignment between laser data and our $C_{map}$ is small. An orthogonal planimetric neighborhood $N$ is considered at each side of the line segments provided as input by the $C_{map}$. The hyper-points of $H_\perp$ having their $(X, Y)$ coordinates included inside the neighborhood $N$ are retained (i.e., subset of points associated to the facades); providing thus disjoint facade clusters (hypothetic facade clusters). $N$ has a sufficiently large tolerance $(N = \pm 1m)$ in order to guarantee that the great majority of the facades are extracted. Finally, each hypothetic facade cluster is retained if the number of included 3D points is above a given threshold (e.g., value fixed here to 500 *pts* according to the estimated 3D point sampling). As can be seen in Fig.3(b), the $C_{map}$ is sometimes incomplete. Since this phenomenon in not too important in our case, missing facade segments of the map have initially been completed by manually chaining existing facade extremities. Besides, we emphasize that the planimetric junction (i.e., boundary) between two joined and coplanar facades is difficult to delineate across a single 3D point cloud of street. Hence, this method presents the advantage that each existing represented facade of the $C_{map}$ is potentially associated to a facade cluster. Results Fig.3(c) clearly show that the goal is reached. Then each extracted cluster can be used for estimating **residential facade delimitations** (stage described in next section). Although this extraction is elementary, it relies on the assumption of the availability of a $C_{map}$ that is well aligned with the point cloud. Therefore, in practice, this direct global extraction cannot be always adopted. In such a case, approximate georeferenced facade segments can alternatively be obtained semi-manually (e.g., procedural selection) or automatically from aerial-based process (e.g., building focusing stage).

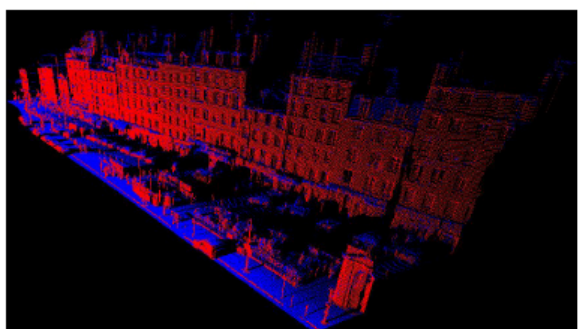

(a) Segmented street point cloud

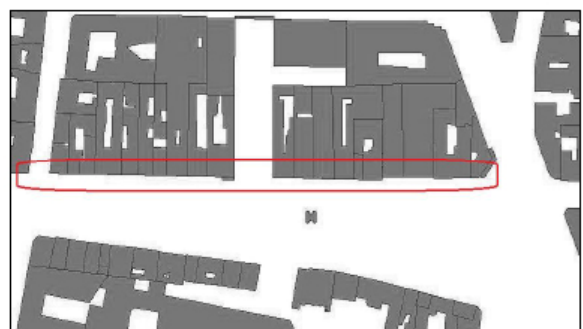

(b) National cadastral map (set of 2D segments)

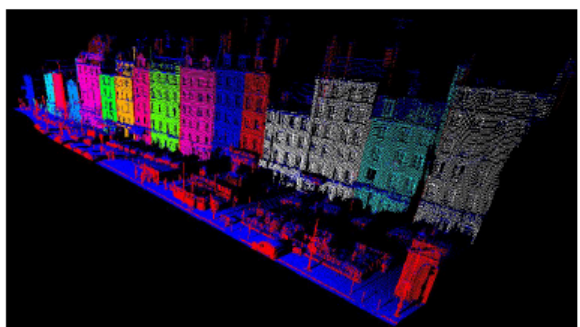

(c) Disjointed facade clusters

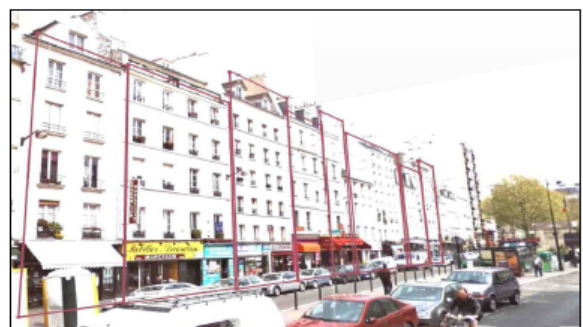

(d) Estimated 3D facade quadrilaterals projected into the corresponding portion of panoramic image

**Figure 3.** Results of the facade cluster individualization and extraction as well as segmentation of the corresponding facade images from laser-based estimation of the 3D facade quadrilaterals and projection.

*3.2.3 Delimiting Vertices of Facade Quadrilateral in Planimetry and Altimetry*

For each retained facade cluster (see Fig.3(c)), the associated dominant vertical facace plane can be estimated by applying a conventional **LSA** technique (Least Square Adjustment technique). Then the orthogonal projection of the extremities of the $C_{map}$ (Fig.3(b)) produces **accurate locations of facade segments** (novel dominant support planes). This **updated facade map is delimited in 3D** by applying the methodology described below. In altimetry, the parked vehicles frequently cause strong occlusions at the bottom of the facade resulting in missing points (see Fig.3(a) and Fig.3(c)). We compensate for this lack of data by retaining the mean altitude of the laser sensor for the set of frames that have voted in each retained facade cluster. The altitude of the facade/ground limit is deduced by subtracting the vehicle height to the mean altitude of the laser sensor for each cluster and adding the height of a standard border pavement. It provides thus a **robust filling gap solutions** in the sense that a local facade bottom delimitation is assigned to each facade plane with a total independence of the importance of occlusions located in front of the facade (see Fig.3(d)). The facade top points can be extracted by retaining for each laser frame the highest point of the retained facade cluster. For each set of retained facade top points, the coordinates are sorted by altitudes and the corresponding median value is determined. For each facade, the **MSD** (Mean Square Deviation) is computed from the heights of the top points. This statistical measure provides thus a score quantifying the altimetric variability and identifies the facade top **LoD** (Level of Detail). If the **MSD** score is high, the maximum altitude amongst points of the cluster is retained in order to ensure a complete facade delineation; else the median altitude is retained to smooth small facade top discontinuities (see Fig.3(d)).

Besides, we emphasize that the generated 3D facade quadrilaterals are derived from the $C_{map}$ but not issued from a direct $C_{map}$ extrusion. The facade quadrilaterals are produced by assembling altimetric and planimetric delimitations (Fig.3(d)) and both of MMS-based laser accuracy and $C_{map}$ global coherency are thus exploited. Besides, if any reliable $C_{map}$ or aerial-based data is available for initialising approximate facade segments, recent automatic techniques can be used for directly estimating accurate and individual facade delimitations by exploiting terrestrial imagery (e.g., [12]). Procedural selection of facade segments can also be envisaged from calibrated terrestrial images.

|  | Facade delimitations |
|---|---|
| Deviation in planimetry | in $(x, y)$ |
| Maximum deviation | $0.913m$ |
| Minimum deviation | $0.014m$ |
| Average deviation | $0.229m$ |

**Table 2.** Comparison between modeling results over 25 facades estimated from terrestrial laser data (best dataset) and their ground-truth models produced by the IGN aerial-based modeling pipeline. The evaluation area deals with a part of the 12[th] district of Paris.

In Table 2, we quantify the geometric accuracy associated with the generated 3D facade quadrilaterals by comparison with a set of corresponding 3D facade quadrilaterals produced by using the IGN aerial-based modeling pipeline (national standard ground truth). This evaluation is conducted over 25 facades in a part of the 12[th] district of Paris. The deviations in planimetry correspond here to the deviations of distance in the $(x, y)$ plane between the extremities of the facade models provided by our ground truth and those of the generated facade models. We mention that the deviation in altimetry (i.e., in altitude) has not been estimated here since our laser dataset was incomplete at facade bottom and/or at facade top in the available ground truth area due to the acquisition specifications. However, we expect that the relative height of facades are estimated with high accuracy due to the accurate laser data.

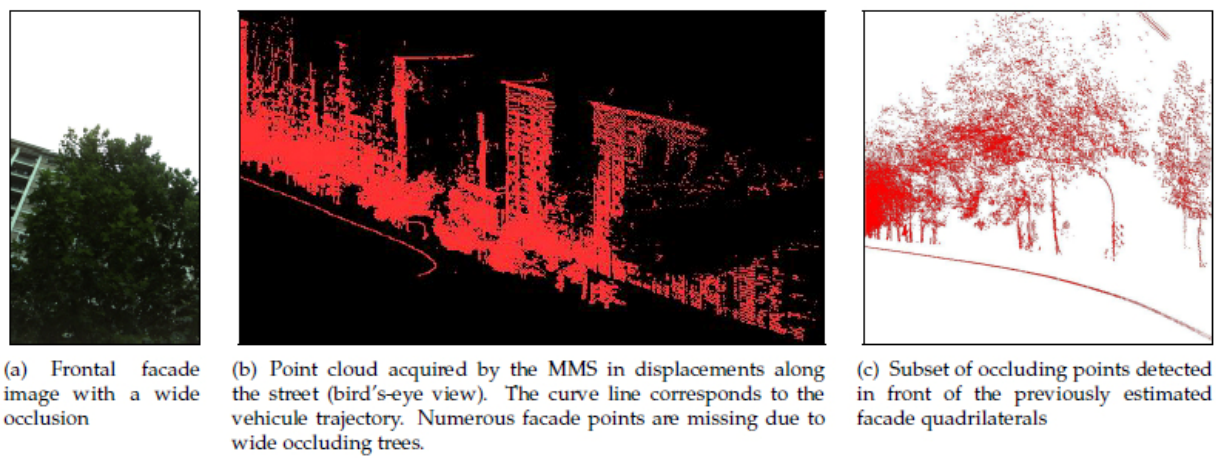

(a) Frontal facade image with a wide occlusion

(b) Point cloud acquired by the MMS in displacements along the street (bird's-eye view). The curve line corresponds to the vehicle trajectory. Numerous facade points are missing due to wide occluding trees.

(c) Subset of occluding points detected in front of the previously estimated facade quadrilaterals

**Figure 4.** Results of the occluding facade point detection by exploiting previously estimated 3D facade quadrilaterals.

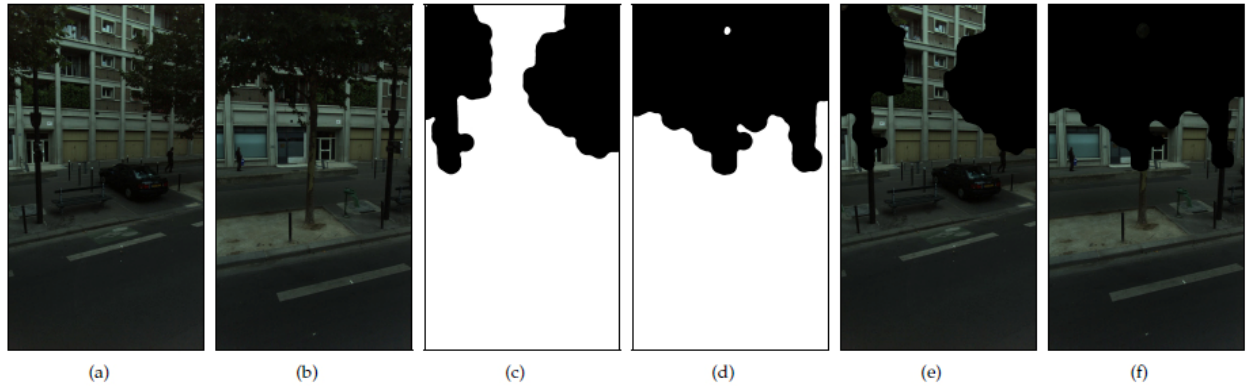

**Figure 5.** Results of the generation of image masks hiding occluding facade objects by fusing detected occluding facade points and multi-view facade images. (a) and (b) are two image samples with projected 3D points. (c) and (d) are associated generated binary masks. (e) and (f) show hidden facade occlusions from the generated masks

In planimetry, the maximum deviation can reach 0.913$m$. This deviation can occur in reason of certain cadastral-based facade segment that can be inaccurate. The average and the minimum deviation respectively are of 0.229$m$ and 0.014$m$. It denotes then that the estimated facade planes are relatively accurate. However, it is worth pointing out the fact that this ground truth is relatively biased since based on the use of a noisy cadastral map and aerial-based noisy data. Furthermore, the accuracy of these delimitations can strongly vary according to some factors such as the GPS information, the intrinsic and extrinsic calibration accuracies of the multi-source sensors, the synchronisation between the positioning systems and the acquisition sensors (vehicle georeferencing), to name a few. Nonetheless, the global topology of the generated wire-frame model appears coherent and the deviation measured in this area are included in a submetric bounded interval. Also, we argue that the generated wire-frame model derived from the terrestrial laser data should be more accurate (ground-based data) since the details between the facade planes are more perceptible (from dense laser data) and since the acquisition is carried out in one single pass. More details regarding this evalutation can be found in [13].

*3.3 Recovering Quasi-real Occlusion-free Facade Textures*

Frequently, facade models are textured by a mapping with frontal or oblique views of buildings. However, these images can be strongly occluded as shown in Fig.4(a). First questions that come to mind in case of an automatic occlusion-free facade texturing can be as follows: **How can we detect facade occlusions in street images? How can we recover occluded facade image parts and reconstruct a complete occlusion-free texture?** We propose in this section a terrestrial-based methodology for answering parts of these open questions even in the case of strong occlusions. The proposed methodology employs the fusion of image and laser data. This facade texturing methodology (for processing strongly occluded facades) is in part inspired by [14] and can be seen as a complementary piece for the powerful approaches described in [15, 16]. We remind that the georeferenced terrestrial laser data have been initially acquired by the MMS in conjunction with a set of calibrated and georeferenced multi-view facade images. The key idea is to exploit the set of optical images acquired by the MMS in order to produce image transfer that is not affected by occluding objects. Since the view point of the MMS will change due to vehicle motion, one can hope that the occluded parts of a given facade in a set of images will be visible in another set. Finally, some inpainting techniques can be applied from the set of images for filling the texture in persistent occluded parts, e.g., non-visible facade parts.

*3.3.1 Detecting Occluding Objects of Urban Facades*

The occluding objects that are observed in the images of facades (e.g., Fig.4(a)) can be detected in the laser data at facade level. This is achieved by identifying that the laser beam is intersected by intermediary occluding objects located between the point of acquisition and the previously estimated facade planes (e.g., street road microstructures described in Tab.(1)). The missing facade laser parts can be qualified as laser shadows and this effect is strongly visible in case of occluding trees since they clearly cause elliptic holes in the facade point cloud (see Figure 4(b)). For this stage, the facade planes have been previously estimated and delimited. We mention that the density of laser points located at the level of facade walls was sufficiently high and points generally were included at facade top (see Fig.4(b)) for guaranteeing the facade extraction and delimitation. Otherwise, the facades quasi-fully occluded (point absent or in low quantity) can be represented by exploiting the remaining facade supports (given by the $C_{map}$) that are previously not matched with facade clusters. They can also be estimated by using existing aerial-based approaches. Since the facade planes and quadrilaterals can be estimated as mentioned earlier, the set of 3D points located in a neighbourhood defined by a facade plane and the vehicle trajectory is retained. In certain cases, the retained set of points can also contain points of the ground according to the sensor orientation. The points of ground can be detected and removed by examining their altitude since the bottom delimitation of the facade has been estimated (sensor-based delimitation). Here, the sensor has been specifically oriented in order to avoid the acquisition of ground points. If the quantity of retained points in the predefined neighbourhood is sufficiently high, then the points are extracted and labelled as occluding points. A result of the occluding point extraction is shown in Fig.4(c). Furthermore, the hyper-points $H_\perp$ previously computed from the accumulation map $A_{map}$ can be used for confirming the presence of facade foreground occluding objects such as previously shown and described in Fig.3(c) and Tab.(1), respectively.

*3.3.2 Generating Masks Hiding the Occluding Objects*

The next stage is to detect the occluding objects in the set of optical images. A set of facade images has been associated to each facade according to a visibility criterion based on the size of the image intersection with the projected 3D facade quadrilaterals (georeferenced). The whole of these images and the detected occluding objects will then be used to the generation of an occlusion-free texture.

More precisely, the detected 3D points that occlude the facade are projected onto the respective set of facade images that have been matched to the facade plane (zoom samples shown Fig.5(a) and Fig.5(b)). We recall that the acquired images are calibrated and the whole of the image and laser data are georeferenced. The detected occluding 3D points are used to generate binary images (masks) with a size identical to the original acquired images (1920 × 1080). Then, these associated binary images (scattered single points) undergo morphological operations.

A morphological dilation is applied onto each projected point in the binary images for amplifying the covering of occlusions and filling gaps. In particular, the structuring element (i.e., kernel) $K_d$ that is employed for the dilation of the binary images B is a large circular kernel (radius of 50 *pixels*) that operates onto the totality of the binary image—scattered single points.

For each detected occluding point, the point can be enhanced into a cube by considering a neighboring, i.e. 3D morphological dilation. The aim of this stage consists in intensifying the density of points in order to produce masks (image masks) that cover more the occluding objects (e.g., at occlusion contours). Visually, we can observe that the coverage of facade occlusions can be enhanced. However this technique is very expensive in computational time in the sense that the number of occluding facade points is multiplied in our case by 9 (e.g.; see results in Fig.6).

A morphological erosion is then applied to the dilated images with a small circular Kernel (radius of 20 *pixels*) in order to reduce the amplified size of the masks at the contours. This step readjusts the masks to the occlusion size. Two examples of the resulting binary masks are shown in Fig.5(c) and Fig.5(d).

Finally, a smoothing is applied to the contours of the masks in order to reduce the visual effect of artificial edges caused by the accumulation of circular Kernel patterns. In this way, the rendering of a mosaicing with facade images by subtracting the masks, i.e. reconstructed textures will be not overly affected at the multiple recovered junctions (see results in Fig.5(e) and Fig.5(f)). Here although the boundaries due to the mask are visible, their appearance is not affected by the kernel's shape. Then, ortho-image of facade images (with masks) are generated. Finally, a raw mosaicing is achieved by overlapping all the processed facade images excluding the masked regions due to occluding objects (Fig. 7(b)).

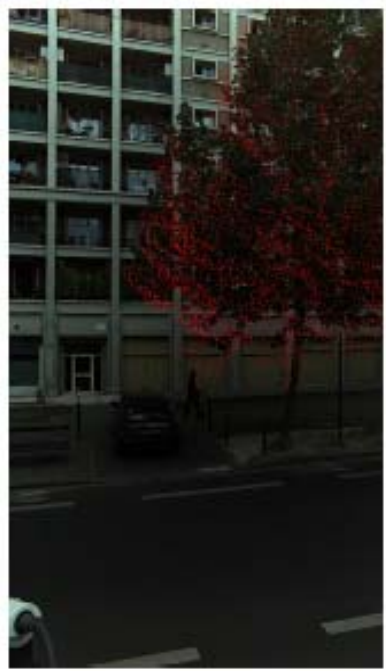

**Figure 6.** Detected 3D occluding facade points and additional 3D points resulting from their 3D dilation that are both projected onto a retained oblique facade image.

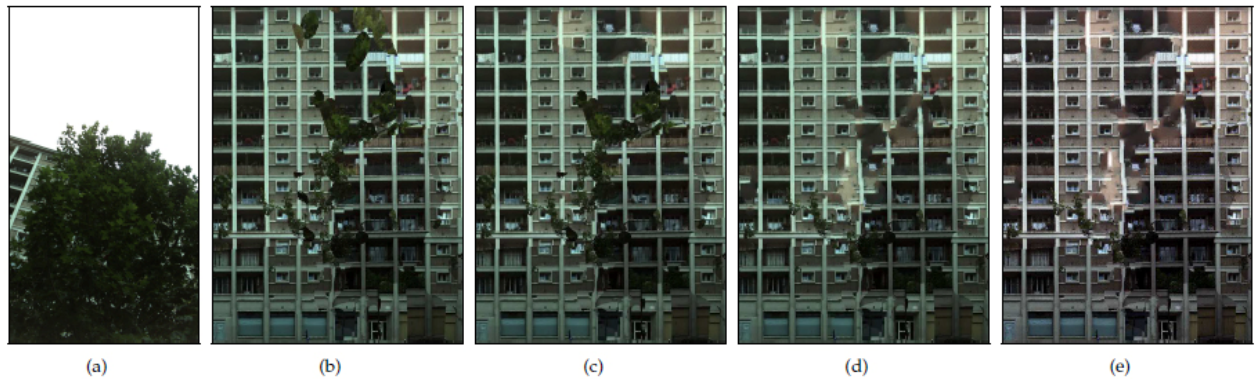

**Figure 7.** Results of the recovered occlusion-free quasi-real texture in the case of a strongly occluded facade. (a) shows a wide frontal facade occlusion. (b) shows the reconstructed textures in multi-view. (c) and (d) show top and central facade regions with resynthesised texture in persisting non-visible facade parts. (e) shows recovered facade true colors.

*3.3.3 Enhancing rendering of reconstructed textures*

As can be seen in Fig.7(b) it happens that the recovered texture is partially incomplete in cases of strong occlusion such as shown in Fig.7(a). In this section, we describe the steps we apply to improve the rendering of this reconstructed texture in order to generate a texture more suitable for visualization. In particular, our interest is essentially focused on remaining occluded texture parts. Inpainting techniques can be semi-manually employed in order to process a subset of textures in cases of extreme occlusion. Notably, techniques such as resynthesizing can be directly applied for the removal of remaining external objects in major critical parts of the reconstructed facade texture. Here, we have used GIMP (GNU Image Manipulation Program) with the Resynthesizer plug-in developed as a part of [17]. This operation requires delineating targeted regions and filling them by using a powerful transfer method based on the remaining surrounding texture. This practical method is a variant of the best-fit methods described in [18] and [19]. Results of resynthesizing are illustrated in Fig.7(c) and Fig.7(d). Moreover, Fig.7(d) presents an unnatural green color tone due to ill-tuned white balances of the cameras or variation of scene illumination during the acquisition. Fig.7(e) shows a result of global adjustment of the color levels by applying an Automatic White Balance (AWB) functionality of GIMP onto the resynthesized texture. Facade true colors are recovered with success. This functionality can also be applied manually by picking master colors. We point out the fact that these semi-manual stages represent minimal interventions and the texture finally produced is quasi-real. Furthermore, we have currently avoided rendering improvements based on the image blurring or image blending in real facade texture parts in spite of remaining imperfections. Indeed, our interest lies in the preservation of the radiometric richness for pixel intensities located in the recovered real facade texture regions (e.g., for image-based post-computations or for zooming).

## 4. Conclusion and Future Works

Facade clusters and quadrilaterals are extracted by fusing a selected cadastral map (facade initialization, global coherence, residential limits, semantic) and laser data (high accuracy, dense sampling). Facade occluding points (detected from the estimated quadrilaterals) are projected onto the associated multi-view images. Occlusion-free facade textures are reconstructed from image masks exploiting fundamental image processing techniques. Aesthetic quality of the produced final textures is enhanced in case of strong occlusions or missing facade regions by using some semi-manual restorations. The obtained results rely on a synergistic use of real urban street laser data, optical images and an existing 2D building map. The proposed technique is able to substantially improve the visualization and the computational modeling.

In particular, since the produced facade maps are georeferenced and retextured without occlusions, this information provides uniform and compatible registration and positioning features for the research in vision-based navigation and localization of street robots along the sidewalks for processed areas. At term, we emphasize that the wide-scale urban facade maps generated from data collected by mapping vehicle sensors could be used for fostering the navigation of street robots; and inversely, the detailed data collected by street robot sensors could be used for completing and extending the wide-scale urban maps with local information.

More expressly, future work will be to vectorize the accumulation map by using image processing algorithms for producing generic facade maps. We also intend to improve the mask geometry (occlusion covering) and radiometry (local equalization) as well as to optimize the number of images used in the mosaicing. Notably, the occlusion covering can be enhanced by using oblique or rotatory sensors. This can also be done by initializing region growing approaches from the facade foreground hyper-points. We also intend to apply mosaicing algorithms (e.g., [20]) to improve the inter-patch alignment and the overall photo-realism of the output as well as automating stages for texture inpainting.

## 5. Acknowledgements

This work was funded by "Institut national de l'information géographique et forestière" (France) and by a Strategic Research Cluster grant (07/SRC/I1168) by Science Foundation Ireland under the National Development Plan. This work was also supported in part by the Spanish Government under the project TIN2010-18856. The authors gratefully acknowledge these supports.